\documentclass{documentstyle/llncs}
\usepackage{caption} 
\usepackage{amsmath,graphicx,wasysym}
\usepackage{amsfonts}
\usepackage{hyperref}
\usepackage{bm}
\usepackage{booktabs}
\usepackage{tabularx}
\usepackage{makecell} 
\usepackage[colorinlistoftodos]{todonotes}
\usepackage{graphicx}
\usepackage{hyperref}
\usepackage{colortbl}
\DeclareMathOperator*{\argmin}{argmin}
\newcommand*{\norm}[1]{\left\lVert#1\right\rVert}

\definecolor{Gray}{gray}{0.9}

\begin{document}

\title{Image Registration and Predictive Modeling: Learning the Metric on the Space of Diffeomorphisms} 

\author{Ayagoz Mussabayeva\inst{1, 3} \and Alexey Kroshnin\inst{1,3} \and Anvar Kurmukov\inst{1,3} \and{Yulia Dodonova}\inst{1} \and Li Shen\inst{4} \and Shan Cong\inst{5} \and Lei Wang\inst{6}  \and Boris A. Gutman\inst{1,2} }

\institute{The Institute for Information Transmission Problems, Moscow, Russia
\and
Department of Biomedical Engineering, Illinois Institute of Technology, Chicago, IL, USA
\and
Higher School of Economics, Moscow, Russia
\and
Department of Biostatistics, Epidemiology and Informatics, University of Pennsylvania, Philadelphia, PA, USA
\and
Indiana University, Indianapolis, IN, USA
\and
Department of Psychiatry and Behavioral Sciences, Northwestern University Feinberg School of Medicine, Chicago, IL, USA
}

%

%
\authorrunning{**} 
%
%


\maketitle              

\begin{abstract}

We present a method for metric optimization in the Large Deformation Diffeomorphic Metric Mapping (LDDMM) framework, by treating the induced Riemannian metric on the space of diffeomorphisms as a kernel in a machine learning context. For simplicity, we choose the kernel Fischer Linear Discriminant Analysis (KLDA) as the framework. Optimizing the kernel parameters in an Expectation-Maximization framework, we define model fidelity via the hinge loss of the decision function. The resulting algorithm optimizes the parameters of the LDDMM norm-inducing differential operator as a solution to a group-wise registration and classification problem. In practice, this may lead to a biology-aware registration, focusing its attention on the predictive task at hand such as identifying the effects of disease. We first tested our algorithm on a synthetic dataset, showing that our parameter selection improves registration quality and classification accuracy. We then tested the algorithm on 3D subcortical shapes from the Schizophrenia cohort Schizconnect. Our Schizpohrenia-Control predictive model showed significant improvement in ROC AUC compared to baseline parameters.

\end{abstract}
\section{Introduction}
\label{sec:intro}

Image registration, and more generally geometric alignment underlies a large number of analyses in medical imaging, particularly neuroimaging. As one of the mainstays of medical image analysis, the problem has been addressed extensively over the last 2+ decades, with several flavors of robust algorithms \cite{Klein_compare_14_registrations}. A number of registration approaches develop an explicit metric space comprised of the geometric objects of interest~--- anatomical shapes, diffusion tensors, images, etc. Prominent among these is the Large Deformation Diffeomorphic Metric Mapping (LDDMM) framework \cite{Original_LDDMM_Beg}.  Instead of treating images as objects of interest directly, LDDMM builds a space of image-matching diffeomorphisms using a Riemannian metric on velocity fields. This metric is induced by a differential operator which at once controls the nature of the metric space and regularizes the registration. 

The structure of such a space~--- a manifold of smooth mappings with well-defined geodesics~--- enables generalizations of several standard statistical analysis methods. These methods adapted to the Riemannian setting has been repeatedly shown to improve their sensitivity and ability to discern population dynamics when compared to projecting the data onto Euclidean domains. Works in this area include computation of the geometric median and metric optimization for robust atlas estimation \cite{Fletcher_Geometric_Median,Zhang_Bayesian_LDDMM_atlas}, time series geodesic regression \cite{Hong_Geodesic_regression}, and principal geodesic analysis \cite{Zhang_Prob_PGA}. 

With the exception of \cite{Zhang_Bayesian_LDDMM_atlas}, in the works above the metric is assumed to be fixed. Further, the metricity and Riemannian inner product with which the LDDMM space is endowed has not been used explicitly in predictive modeling up to now. In this work, we strive for two complementary aims: (1) to exploit the Riemannian metric on registration-defining velocities as a kernel in a classification task and (2) to optimize the metric to improve classification. We follow an Expectation-Maximization (EM) approach similar to \cite{Zhang_Bayesian_LDDMM_atlas}, alternating between minimizing image misalignment for kernel estimation, and optimizing model quality over the kernel parameters. In this work, we choose the kernel Fischer linear discriminant classifier for simplicity, though other predictive models are admissible in our framework as well. It is our hope that by explicit tuning the diffeomorphism metric to questions of biological interest, the carefully crafted manifold properties of LDDMM will gain greater practical utility. 

Our experiments consist of synthetic 2-dimensional shape classification, as well as classifying hippocampal shapes extracted from brain MRI of the Schizconnect schizophrenia study. In both cases, the classification accuracy and ROC area under the curve (AUC) improved significantly compared to default baseline kernel parameters.

\section{Methods}

\subsection{Metric on diffeomorphisms}
\label{sec:lddmm}

The Large Deformation Diffeomorphic Metric Mapping (LDDMM) was first introduced in \cite{Original_LDDMM_Beg}. The goal of the registration is to compute a diffeomorphism $\phi\colon \Omega \rightarrow \Omega$, where the $\Omega$ is the image domain. The diffeomorphism $\phi$ is generated by the flow of a time-dependent velocity vector field $v$, defined as follows:  
\begin{equation}  \label{eq:diff_velocity}  
\begin{gathered}
\frac{\partial \phi(t, x)}{\partial t} = v(t, \phi(t, x)),\\
\phi(0, x) = id,
\end{gathered}
\end{equation}
where $id$ is the identity transformation: $id(x) = x$, $\forall x \in \Omega$.
This equation gives a path $\phi_t\colon \Omega \rightarrow \Omega$, $t \in [0, 1]$, in the space of diffeomorphisms. Estimation of the optimal diffeomorphism via the basic variational problem in the space of smooth velocity fields $V$ on $\Omega$ takes the following form, constrained by \ref{eq:diff_velocity}:  
\begin{equation}  \label{eq:opt_LDDMM}  
v^{*} = \argmin_{v} \left(\int\limits_0^1 \norm{v_t}^2_L \,d t + \frac{1}{\sigma^2} \norm{I_0 \circ \phi - I_1}^2\right).
\end{equation}
The required smoothness is enforced by defining the norm $\norm{\cdot}_L$ on the space $V$ of smooth velocity vector fields through a Riemannian metric $L$.

The Riemannian metric $L$ should be naturally defined by the geometric structure of the domain. The inner product $\norm{v}^2_L = \langle L v, v \rangle$ can also be thought of as a metric between images, i.e.\ the minimal diffeomorphism required to transform the appearance of $I_0$ to be as similar as possible to $I_1$. Since the diffeomorphism space is a Lie Group with respect to eq.\ref{eq:diff_velocity}, the Riemannian metric defined suggests a right-invariance property. The original LDDMM work \cite{Original_LDDMM_Beg} defines $L$ as a smooth differential self-adjoint operator $L =(\alpha \Delta + \beta E)$, where E is identity operator. Here, we choose to use an $L$ based on the biharmonic operator $L =(\alpha \Delta + \beta E)^2$, as e.g. in \cite{Mang2016DistributedMemoryLD}. The parameters $(\alpha, \beta)$ correspond to convexity and normalization terms, respectively. These parameters significantly affect the quality of the registration. It is not obvious how to select $\alpha$ and $\beta$, though they effectively define the geometric structure of the primal domain. Indeed, these are the parameters we optimize in our EM scheme below. 


\subsection{Predictive Model}
\label{sec:klda}

Consider a standard binary classification problem: given a sample $\bigl(x_i, y(x_i)\bigr)_{i = 1}^n$, where $y(x_i) \in \{1, -1\}$ is a class label, find a classification function $\hat{y}$ approximating the true one $y$. 
One  of the standard linear techniques in statistical data analysis is the Fisher's linear discriminant analysis (LDA). The kernel Fisher Discriminant Analysis (KLDA) introduced in \cite{KLDA_Original} is a generalization of the classical LDA. There are several approaches to derive more general class separability criteria. KLDA derives a linear classification in the embedding feature space (RKHS\cite{aronszajn1950theory}) induced by a kernel $k$, what corresponds to a non-linear decision function in the original, or ``input'' space. The main idea of LDA is to find a one-dimensional projection $w$ in the feature space that maximizes the between-class variance while minimizing the within-class variance. KLDA seeks an analogous projection in the embedding space, where means $(M_z)$ and covariance matrices $(\Sigma_z)$ for each class $z \in \{-1, 1\}$ are computed. The (K)LDA cost function takes the following quadratic rational form:
\begin{equation}  \label{eq:opt_KLDA}  
J(w) = \frac{w^T (M_1 - M_{-1}) (M_1 - M_{-1})^T w}{w^T (\Sigma_{-1} + \Sigma_1) w} = \frac{w^{T} M w}{w^T N w},
\end{equation}
where
\begin{gather*}
	(M_z)_i = \frac{1}{n_z} \sum_{x_\ell : y(x_\ell) = z} k(x_i, x_\ell),\\
    (\Sigma_z)_{i, j} = \frac{1}{n_z} \sum_{x_\ell : y(x_\ell) = z} k(x_i, x_\ell) k(x_j, x_\ell) - (M_z)_i (M_z)_j.
\end{gather*}
Here $n_z$ is a number of objects from class $z$ in the sample.

The solution of the problem $J(w) \rightarrow \min$ is known to be $\hat{w} = N^{-1} (M_1 - M_{-1})$. The decision function for a new observation $x$ is based on the projected distance to the training sample means, $w^T (M_z - x)$. 
The M-step in an EM formulation requires a differentiable measure of model quality which in our case is the accuracy of classification. The more common approach is to formulate a probabilistic model which leads to a log-likelihood optimization. Such an approach is used e.g.\ in \cite{Zhang_Bayesian_LDDMM_atlas}. In our case, this can be done by modeling the classifier's output with a parametric distribution. 

However, we found that such a formulation using the sigmoid distribution function leads to an unstable solution. Instead, we propose to use the hinge loss defined for KLDA as 
\begin{equation}  \label{eq:hinge_loss}  
\begin{gathered}
h(x', z) = \max\{0,\, 1 - z \, y(x')\}\\
y(x') = \sum\limits_{i = 1}^n w_i \left(k(x_i, x') - \frac{(M_1)_i + (M_{-1})_i}{2}\right),
\end{gathered}
\end{equation}
where $z \in \{-1, 1\}$ is a true label for the new observation $x'$ and $k(x_i, x')$ is the inner product (i.e.\ the kernel) between $x'$ and the training observation $x_i$. While both hinge loss and log-likelihood formulations eventually lead to some locally optimal solutions on simple problems, such as our synthetic dataset, the former exhibits greater stability. For the hippocampal data, only hinge loss minimization leads to a stable solution.

\subsection{Learning the diffeomorphic metric}
\label{sec:reg+ml}

The main goal of this work is to use the registration-derived metric to classify images. Let us denote the Riemannian metric by $K_L(\alpha, \beta) = \langle Lv, v \rangle$. In practice, $\beta$ plays an insignificant role and can be fixed, as multiplication of the velocity by a constant does not change the optimization problem in LDDMM. We focus on optimizing $\alpha$, fixing $\beta = 1$ as a normalization term. 

We optimize $\alpha$ in the EM framework as follows.

$\bullet$ E-step:

Register each pair of images in our training sample optimizing equation \ref{eq:opt_LDDMM} to derive $K_L(x_i, x_j)$. Define $K(x_i, x_j) = \exp\{-\gamma K_{L}(x_i, x_j)\}$ 
and apply KLDA using $K(x_i, x_j)$. The parameter gamma is estimated by grid search to make a computation easier, but it can be also estimated by gradient descent. Estimate the hinge loss \ref{eq:hinge_loss} given a fixed $\alpha$. 

$\bullet$ M-step:

Minimize the hinge loss \ref{eq:hinge_loss} with respect to $\alpha$.

The primary computational challenge above is in the M-step. Though the decision function is non-convex with respect to $\alpha$, we seek a local minimum via gradient descent. We give the gradient direction with respect to $\theta = (\alpha,\beta)$ below, keeping in mind that $\beta$ is fixed.

\begin{equation}  \label{eq:derivative_hinge_loss}  
\begin{gathered}
\frac{d h(x',z)}{d \theta} = \begin{cases}
-z \frac{d y(x')}{d \theta},& \text{if } z y(x') < 1,\\
0,& \text{otherwise}.
\end{cases}
\end{gathered}
\end{equation}

Using the matrix notation $y(x') = w^T \bigl(k(x') - (M_1 + M_{-1}) / 2\bigr)$ with $k_i(x') = K(x_i, x')$ one can obtain
\begin{equation}  \label{eq:derivative_hinge_loss}  
\begin{gathered}
\frac{d y(x')}{d \theta} = \left(k(x') - \frac{M_1 + M_{-1}}{2}\right)^T \frac{d w}{d \theta} + w^T \left(\frac{d k(x')}{d \theta} - \frac{1}{2} \frac{d (M_1 + M_{-1})}{d \theta}\right),\\
d w = - N^{-1} (d N) N^{-1} (M_1 - M_{-1}) + N^{-1} (d M_1 - d M_{-1}),\\
\frac{d K(x_i, x')}{d \theta} = -\gamma K(x_i, x') \frac{d K_{L}(x_i, x')}{d \theta}\\
\frac{d K_{L}(x_i, x')}{d \theta} = \begin{pmatrix}
           2 \bigl\langle (\alpha \Delta^2 + \beta \Delta) v_{x, x'}, v_{x, x'}\bigr\rangle,\\
           2 \bigl\langle (\alpha \Delta + \beta E) v_{x, x'}, v_{x, x'}\bigr\rangle. 
         \end{pmatrix}
\end{gathered}
\end{equation}

The resulting algorithm requires $n \times (n - 1)$ registrations at each EM step to train, and $n$ registrations to a new image from each of $n$ images in training sample to apply. 
\section{Experiments}
\label{sec:experiments}

To derive a baseline set of metrics between pairs of images, we selected $\alpha$ to maximize mutual information between registered images. This metric was then used to define the kernel in the KLDA classifier, the results of which we used as a baseline accuracy for our proposed method. 
\begin{figure} [ht]
\begin{center}
\begin{tabular}{c} 
   \includegraphics[height=6.5cm]{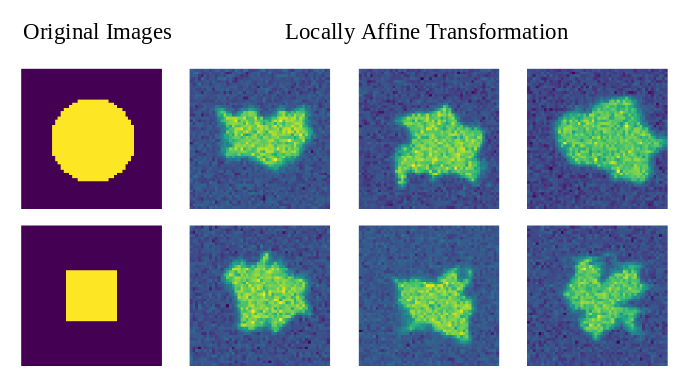}
   \end{tabular}
   \end{center}
   \caption[example] 
   { \label{fig:synthetic_data_ex} 
\textbf{Synthetic data generation} 
}
\end{figure} 

\begin{figure}[!htb]
  \centering
  \begin{minipage}[b]{0.49\textwidth}
    \includegraphics[width=\textwidth]{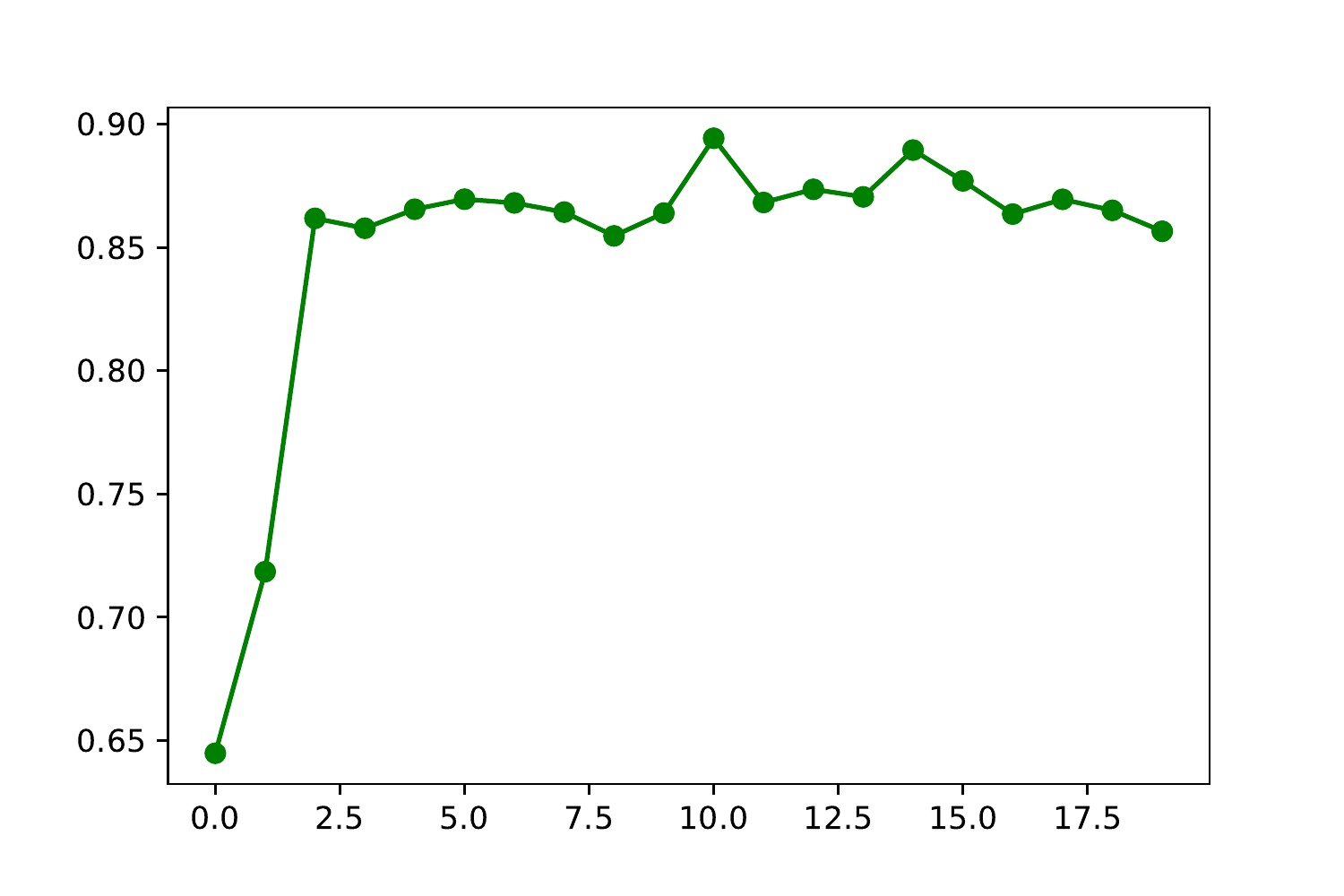}
    \end{minipage}
  \hfill
  \begin{minipage}[b]{0.49\textwidth}
    \includegraphics[width=\textwidth]{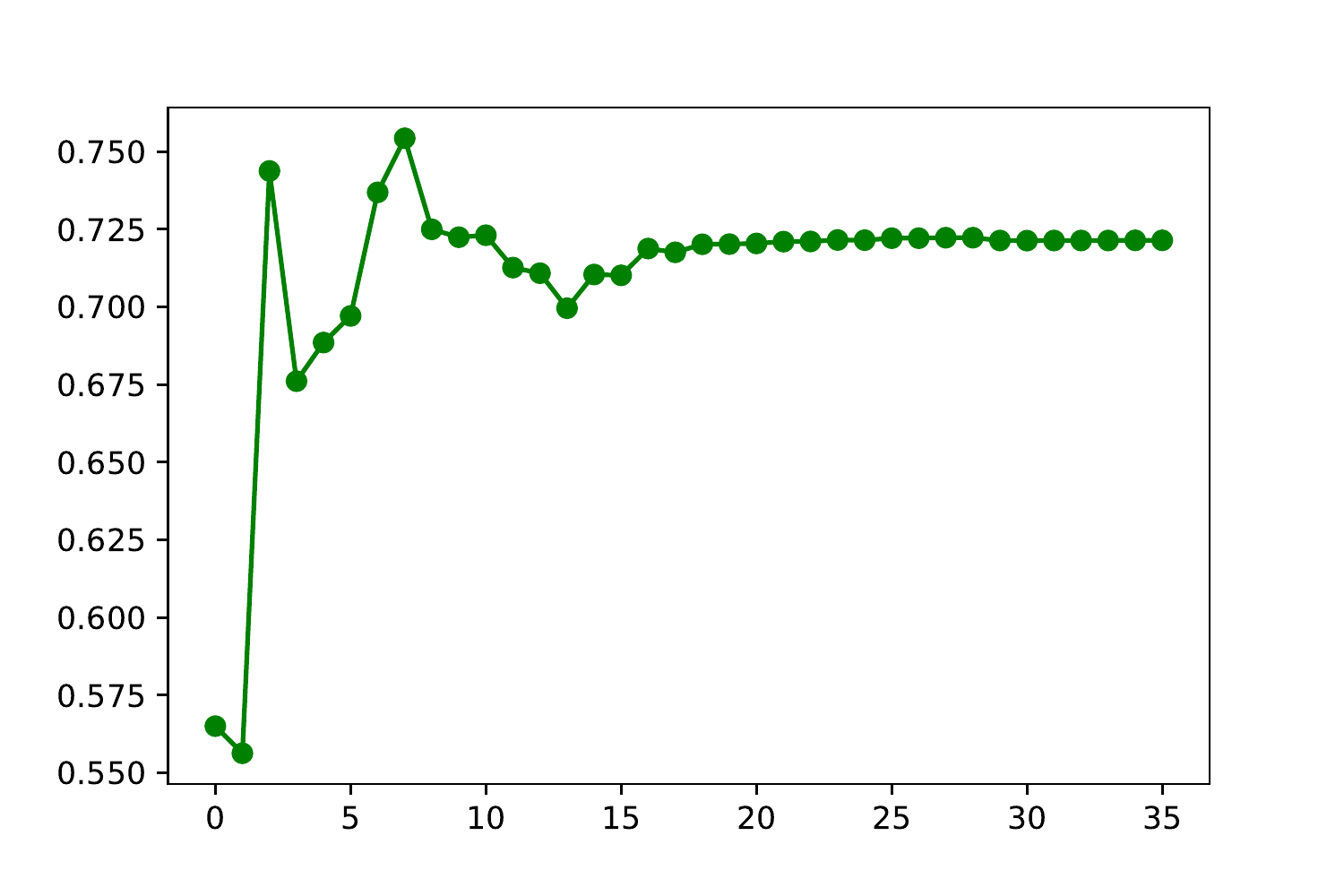}
    \end{minipage}
  \caption 
   { \label{fig:roc_auc} ROC area under the curve vs. EM iterations on \textbf{(left) synthetic data} and \textbf{(right) hippocampal shape} 
}. 
\end{figure}

Our initial experiments were based on 100 images of rectangles, and 100 images of ellipses, each generated with a random locally affine deformation sufficiently noisy to obscure the original class of the image to the naked eye \textbf{(Figure \ref{fig:synthetic_data_ex})}. 
Using 50 deformed ellipses and 50 deformed rectangles as a training dataset, we optimized $\alpha$ until hinge loss convergence. \textbf{Figure  \ref{fig:roc_auc}) (left)} 
shows that EM converges stably after several iterations based on ROC AUC. The final model, chosen based on the best training ROC AUC, performed nearly as well on the synthetic test dataset: ROC AUC = 0.84. 

Our 3D hippocampal shape sample was derived from the SchizConnect brain MRI dataset \cite{schizconnect}. We used right hippocampal segmentations extracted with FreeSurfer \cite{fischl2012freesurfer} from 227 Schizophrenia (SCZ) patients and 496 controls (CTL). All shapes were affinely registered to the ENIGMA hippocampal shape atlas \cite{ShapeNatureComm}, and their binary mask was computed from the transformed mesh model.

We again used 100 training examples (50 CTL, 50 SCZ) in all our experiments below, using the remaining sample as a test dataset. To derive baseline results to compare with our algorithm's performance on hippocampal shapes, we constructed two additional discriminative models. 

(1) A logistic regression model simply using the vectorized binary mask. No spatial information is used in this model.

(2) A KLDA model constructed using LDDMM metrics optimized for registration quality. 

ROC AUC scores for the three models are shown in \ref{tb:result}.

\begin{figure} [ht]
\begin{center}
\begin{tabular}{c} 
   \includegraphics[height=10cm]{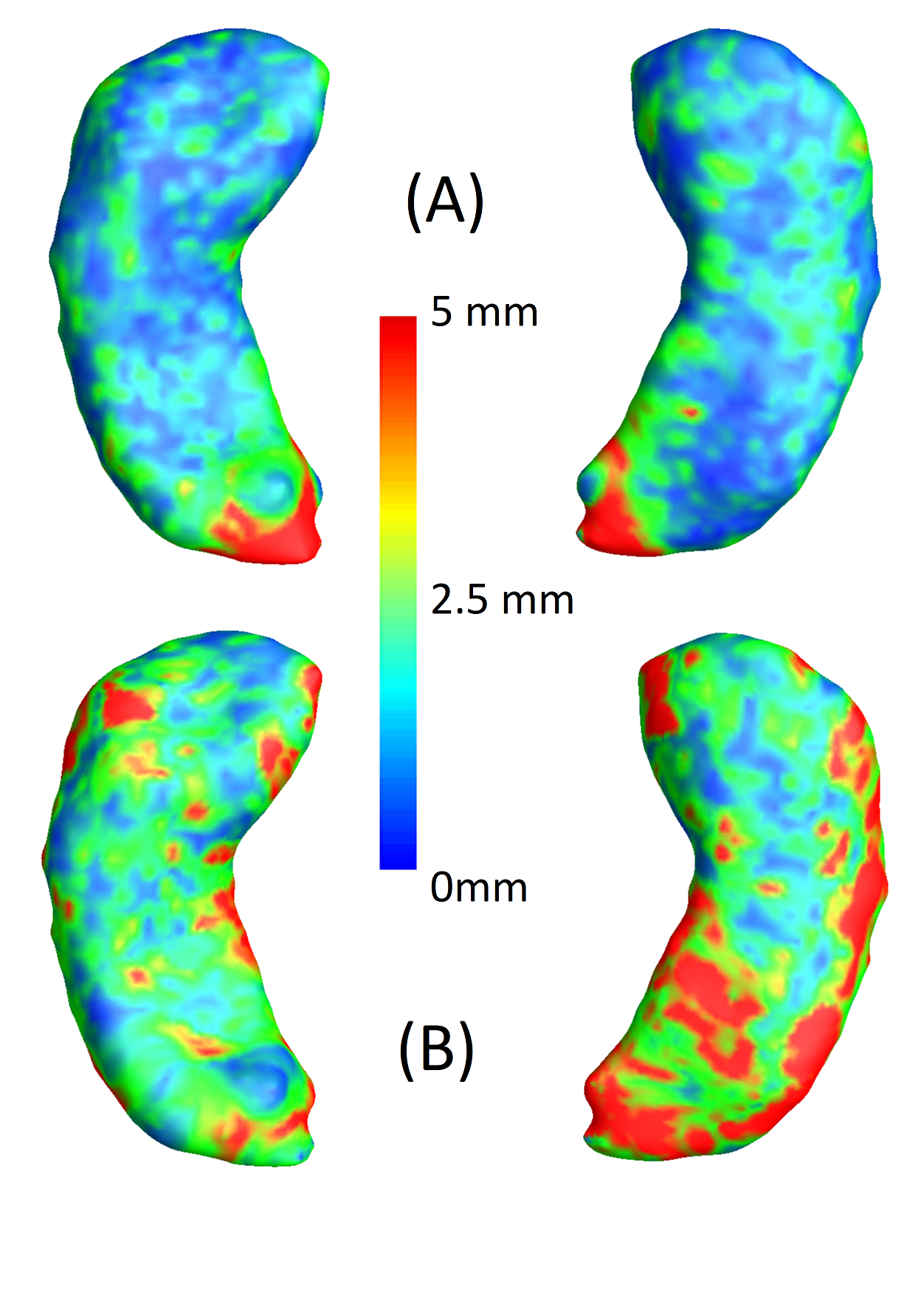}
   \end{tabular}
   \end{center}
   \caption[example] 
   { \label{fig:momenta} 
\textbf{Mean momentum difference between Schizophrenics and healthy subjects}, using (A) a classification-optimized metric and (B) a metric optimized for pairwise mutual information. The effect in the latter is diffuse, while the classification-aware metric focuses on the hippocampal tail. 
}
\end{figure}

\begin{table}[h]
\centering
\begin{tabular}{|p{1.6cm}|p{2.8cm}|p{2.3cm}|p{4.2cm}|}
\hline
{} & Logistic Regression &  Maximum MI  & Optimized LDDMM-kernel\\ \hline
ROC AUC  &0.36$\pm$0.02 & 0.72 $\pm$ 0.06  & 0.75 $\pm$ 0.06 \\\hline
\end{tabular}
\caption{ROC AUC scores for three models}
\label{tb:result}
\end{table}
As expected, ignoring spatial information leads to significant drop in performance. It is also encouraging to see improvement in the classification accuracy when the LDDMM metric is optimized for this explicitly. The stability of the EM algorithm trained on hippocampal shapes is comparable to  stability when synthetic data, as seen in \textbf{Figure  \ref{fig:roc_auc}) (right)}. To visualize the difference in the kernel-based models, we project the mean difference between SCZ subjects and controls in the scalar momenta defining the registration velocity fields \cite{Mang2016DistributedMemoryLD}, as seen in \textbf{Figure \ref{fig:momenta}}. 

\section{Conclusion}

We have presented a method to optimize registration parameters for improved classification performance. Method exploits the geodesic distance on the space of diffeomorphisms as an image similarity measure to be learned in the fashion of traditional metric learning \cite{MetricLearningSurvey}. Our aim in this work was twofold: 1. to show that the metricity of a high dimensional space of geometric objects can be successfully used to improve predictive modeling, and (2) to suggest a means of making the sophisticated mathematical machinery of constructions such a LDDMM more useful in medical imaging practice. As a first attempt, we believe this work shows progress towards both goals. A stable LDDMM metric optimization is devised, and classification accuracy in our real-world application is indeed improved. The main drawback is the significant computational burden, as $N \times N$ training registrations are required. One approach to alleviate this problem is to lift the classification problem onto the tangent space at identity, thus requiring only $N$ training registrations to an atlas, similar to \cite{Zhang_Bayesian_LDDMM_atlas}. Other generalizations of the idea presented here are possible both in LDDMM and other metric frameworks. We hope our work will inspire these generalizations to be developed. 

\label{sec:conclu}

\section{Acknowledgements}

This work was funded in part by the Russian Science Foundation grant 17-11-01390.

\label{sec:ack}
%
%
\bibliography{biblio/shape_refs}
\bibliographystyle{biblio/splncs}


\end{document}